\documentclass[journal]{IEEEtran}

\usepackage{amssymb}
\usepackage{amsmath}
\usepackage[dvipdfmx]{graphicx} \usepackage{bmpsize}
\usepackage[hidelinks]{hyperref}
\usepackage{ragged2e}

\def\ie{i.e.,~}
\def\eg{e.g.,~}
\def\etal{et al.}

\usepackage[noadjust]{cite}

\usepackage{multirow}
\usepackage[usenames]{color}
\definecolor{purple}{rgb}{1,0,1}
\definecolor{lightgray}{rgb}{0.7,0.7,0.7}

\newcommand{\term}[1]{\emph{#1}}

\begin{document}

\title{Fully Convolutional Neural Networks to Detect Clinical Dermoscopic Features}

\author{Jeremy Kawahara and Ghassan Hamarneh,~\IEEEmembership{Senior Member,~IEEE}
\thanks{Manuscript received December 5, 2017; revised March 28, 2018;
accepted April 20, 2018. Date of publication May 1, 2018. This work was supported by the Natural Sciences and Engineering Research Council of Canada. (\emph{Corresponding author: Jeremy Kawahara.})}
\thanks{J. Kawahara and G. Hamarneh are with the School of Computing Science, Simon Fraser University, Burnaby BC V5A
1S6, Canada
\mbox{(e-mail: jkawahar@sfu.ca}; hamarneh@sfu.ca).}
\thanks{Digital Object Identifier \href{https://doi.org/10.1109/JBHI.2018.2831680}{10.1109/JBHI.2018.2831680}}
}

\markboth{IEEE Journal of Biomedical and Health Informatics, Vol. 23, No. 2, March 2019}%
{Kawahara \MakeLowercase{\textit{et al.}}: Fully Convolutional Neural Networks to Detect Clinical Dermoscopic Features}

\IEEEoverridecommandlockouts
\IEEEpubid{\begin{minipage}[t]{\textwidth}\ \\[10pt]
		\centering\footnotesize{\copyright 2018 IEEE. Personal use of this material is permitted. Permission from IEEE must be obtained for all other uses, in any current or future media, including reprinting/republishing this material for advertising or promotional purposes, creating new collective works, for resale or redistribution to servers or lists, or reuse of any copyrighted component of this work in other works}
	\end{minipage}} 

\maketitle              

\begin{abstract}
The presence of certain clinical dermoscopic features within a skin lesion may indicate melanoma, and automatically detecting these features may lead to more quantitative and reproducible diagnoses. We reformulate the task of classifying clinical dermoscopic features within superpixels as a segmentation problem, and propose a fully convolutional neural network to detect clinical dermoscopic features from dermoscopy skin lesion images. Our neural network architecture uses interpolated feature maps from several intermediate network layers, and addresses imbalanced labels by minimizing a negative multi-label Dice-F$_1$ score, where the score is computed across the mini-batch for each label. Our approach ranked first place in the \emph{2017 ISIC-ISBI Part 2: Dermoscopic Feature Classification Task} challenge over both the provided validation and test datasets, achieving a 0.895\% area under the receiver operator characteristic curve score. We show how simple baseline models can outrank state-of-the-art approaches when using the official metrics of the challenge, and propose to use a fuzzy Jaccard Index that ignores the empty set (\ie masks devoid of positive pixels) when ranking models. Our results suggest that (i) the classification of clinical dermoscopic features can be effectively approached as a segmentation problem, and (ii) the current metrics used to rank models may not well capture the efficacy of the model. We plan to make our trained model and code publicly available.
\end{abstract}


\section{Introduction}

\IEEEPARstart{I}{n} order to distinguish melanoma from benign lesions, dermatologists often rely on using melanoma-specific image cues to aid in their diagnosis. Dermoscopy images, which are captured with a dermatoscope, offer a magnified view of the skin lesion and allow dermatologists to visualize structures within the lesion that may indicate melanoma~\cite{Korotkov2012}. For example, the 7-point checklist~\cite{Argenziano1998} is a scoring system that checks for the presence of visual cues (\eg streaks) in dermoscopy images, and assigns a numerical score that, if exceeded, may indicate melanoma. This helps give dermatologists an objective criteria on which to base their diagnosis.

\subsubsection*{Detecting Dermoscopic Features}
Many groups have studied how to detect and classify clinical dermoscopic features from dermoscopy. Celebi~\etal~\cite{Celebi2008} detected the blue-whitish veil in dermoscopy images. They formed a feature vector using colour and texture based features from patches of pixels, and used a decision tree to classify the patch. Sadeghi \etal~\cite{Sadeghi2013} proposed geometric, structural, orientation, and chromatic features to capture the properties of streaks. Combined with colour and texture based features, they classified absent, regular, and irregular streaks.  Mirzaalian~\etal~\cite{Mirzaalian2012a} modeled the tubular properties of streaks with a Hessian based tubular filter. They computed a feature vector by measuring the detected flux through multiple iso-distance contours to the lesion’s boundary, and trained a support vector machine (SVM) classifier to classify absent, regular, or irregular streaks. Barata~\etal~\cite{Barata2012} proposed using directional filters in dermoscopy images to detect the presence of pigment networks. They formed feature vectors used for classification based on the density and distribution properties of the detected pigment networks.

\subsubsection*{Deep Learning to Segment and Classify Skin Lesions}
Previous work has shown convolutional neural networks (CNNs) to be useful for both skin
lesion segmentation and classification  tasks~\cite{Codella2015,Codella2017,Kawahara2016a,Kawahara2016,Romero-Lopez2017,Yu2017}. CNNs have stacked layers of convolution filters with, commonly, millions of free parameters (also called weights) that learn to represent the data at different levels of abstraction~\cite{LeCun2015a}. These free parameters are often learned through a training process where example images and their corresponding labels (e.g., diagnoses or segmentation masks) are used to update the CNN's free parameters such that the network learns to produce outputs that match the labels. In order to learn free parameters that give a useful abstraction of the data, CNNs often are trained on large datasets of images. As existing skin datasets are relatively small, a common approach~\cite{Codella2015,Codella2017,Kawahara2016a,Kawahara2016,Romero-Lopez2017,Yu2017} is to use the parameters of a CNN already trained over a larger dataset~\cite{Russakovsky2014}. This leverages the useful data abstractions learned over larger datasets for smaller datasets.

\subsubsection*{S\o{}rensen-Dice-F$_1$ Score as a Loss Function}
Training a CNN typically requires minimizing a loss function. As the network model parameters are updated to minimize the loss, the choice of the loss function influences the resulting trained model. The S\o{}rensen-Dice coefficient or F$_1$ score has been proposed as a loss function for imbalanced datasets~\cite{Pastor-Pellicer2013,Milletari2016,Sudre2017}. We note that the S\o{}rensen-Dice coefficient and the F$_1$ score are equivalent (discussed in Section~\ref{isic:sec:loss}). Pastor-Pellicer~\etal~\cite{Pastor-Pellicer2013} proposed the negative F$_1$ score as a loss function for neural networks in order to clean and enhance ancient document images. Milletari \etal~\cite{Milletari2016} proposed using the S\o{}rensen-Dice coefficient as the loss function for a neural network designed for volumetric segmentation. Sudre \etal~\cite{Sudre2017} proposed using the S\o{}rensen-Dice coefficient weighted by the size of the object within the image as the neural network loss function for 2D and 3D segmentation.

\begin{figure*}
\begin{center}
\includegraphics[width=0.99\linewidth]{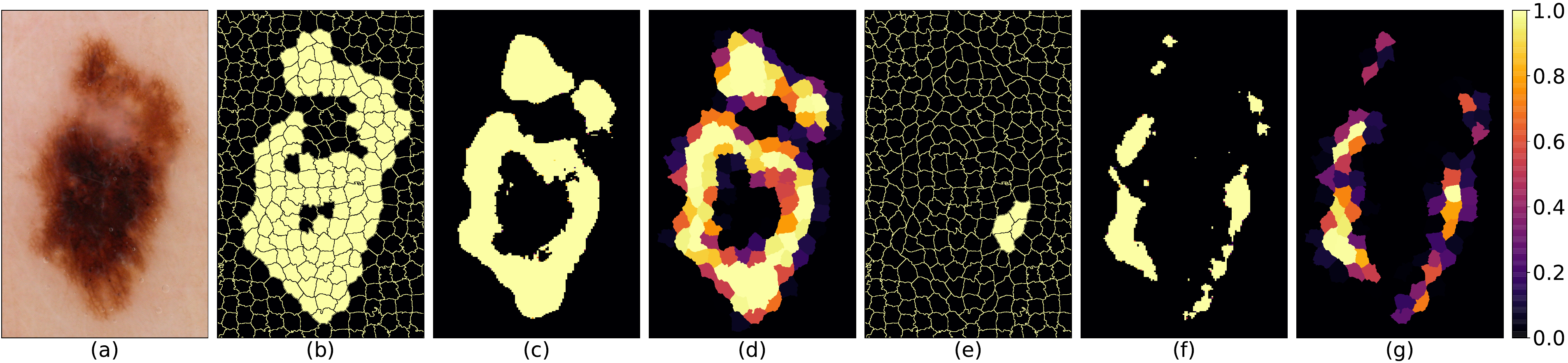}
\end{center}
\caption{Superpixels to segmentations, and segmentations to superpixels. (\emph{a}) The original image. Expertly annotated (\emph{b}) pigment-network and (\emph{e}) streak superpixels converted to binary segmentations, overlaid with superpixels. Pixel-wise (\emph{c}) pigment-network and (\emph{f}) streaks CNN predictions. CNN predictions converted to (\emph{d}) pigment-network and (\emph{g}) streak superpixels. Images shown here are cropped around the lesion for visualization purposes.}
\label{fig:superpixelseg}
\end{figure*}

\subsubsection*{Skin Lesion Datasets and Competitions}
Korotkov \etal~\cite{Korotkov2012} noted that one of the major limitations of computerized skin lesion analysis research is the lack of standardized skin lesion datasets, and that the ``creation of such a dataset is of utmost importance for future development of this field". Fortunately, since this review, new skin lesion datasets have become available such as DermoFit~\cite{Ballerini2013}, and PH$^2$~\cite{Mendonca2013}. More recently, the International Skin Imaging Collaboration (ISIC), in conjunction with the IEEE International Symposium on Biomedical Imaging (ISBI), began hosting a skin lesion analysis competition~\cite{Gutman2016,Codella2017a}. In addition to providing a standardized dataset, this public competition offers standard evaluation procedures and metrics in order to benchmark lesion segmentation, dermoscopic feature detection, and lesion classification approaches. In this work, we focus on \emph{Part 2: Dermoscopic Feature Classification Task} of the 2017 ISIC-ISBI challenge~\cite{Codella2017a}. This task involves classifying superpixels that may contain a specific clinical dermoscopic feature.

\subsubsection*{Contributions}
In this work, we detail our proposed approach that reformulates the superpixel classification task as a segmentation problem, and finetunes a pretrained CNN to detect pixels that contain the studied clinical features. Our CNN architecture is modified for semantic segmentation, and is trained to minimize a negative multi-label fuzzy S\o{}rensen-Dice-F1 score, where the score is computed over partitions of the mini-batch. This approach ranked first place in the \emph{2017 ISIC-ISBI Part 2} task~\cite{Codella2017a}, which used the area under the receiver operator characteristic curve (AUROC) to evaluate submissions. We discuss the limitations of the metrics used to rank the challenge entries, and show two simple baseline methods that empirically outperform all entries when ranked by the current and past challenge metrics. We propose to use a fuzzy Jaccard Index that ignores the empty set (\ie when neither predicted nor ground contain positive values) to rank model performance, rather than AUROC. We plan to publicly release our trained model along with the code used to create and train the model.

\section{Methods}
Given a dermoscopy image $x$, and a corresponding superpixel labelling mask $s$, our task is to predict the set of labels $l$ that belong to each superpixel. The $i$-th label $l_i$ assigns the superpixel $s_i$ the following $K$ potentially overlapping dermoscopic features: \term{pigment network}; \term{negative network}; \term{milia-like cysts}; and \term{streaks}. These are represented as binary vectors of length $K=4$. For example, $l_i = [1,0,0,1]$ indicates that the $i$-th superpixel contains both a \term{pigment network} and a \emph{streaks} dermoscopic feature. 

\subsubsection*{Motivations to Segment Instead of Label Superpixels}
While labelling superpixels is a convenient way to gather ground truth data from human clinicians as it avoids a detailed per-pixel labelling, individual superpixel labelling is less desirable for machine classification tasks for the following two reasons. Firstly, by considering each superpixel individually, the machine classifies based only on the local context available within a superpixel, and ignores surrounding context such as location relative to the entire lesion (\eg dermoscopic features commonly occur within or near the border of the lesion). Secondly, many state-of-the-art approaches for classification rely on a deep learning framework~\cite{Russakovsky2014}. Classifying individual superpixels within a deep learning framework is challenging, as typical deep learning frameworks expect a fixed sized rectangular input, whereas, individual superpixels are of varying size and have non-rectangular shapes. Further, converting to a more conventional deep learning approach allows us to take advantage of neural networks pretrained over larger datasets.

\subsection{Superpixels to Segmentations} 
As previously motivated, rather than treating this as a superpixel classification problem, we instead model this as a multi-label segmentation task. We convert the superpixels $s$ and corresponding labels $l$ into a 3D volume $m \in \mathbb{Z}^{K \times W \times H}$, where $K$ indicates the number of labels, and the width $W$ and height $H$ correspond to the spatial dimensions of the input image $x$ (Fig.~\ref{fig:superpixelseg}\emph{a}). Specifically, we assign each element within $m$ a binary label $l_{ik}$ to indicate the presence or absence of the $k$-th dermoscopic feature at a particular element $m_{kwh}$,
\begin{equation}
(w,h) \in s_i \implies (m_{kwh} = l_{ik})
\label{isic:eq:super_to_seg}
\end{equation}
where $l_{ik}$ represents the $k$-th label for the $i$-th superpixel, and the superpixel $s_i$ is composed of $(w,h)$ spatial locations that index into the spatial locations of $m$. This representation captures the spatial dependencies among superpixels, and allows us to efficiently leverage pretrained CNNs.

\begin{figure*}
\begin{center}
\includegraphics[width=0.98\linewidth]{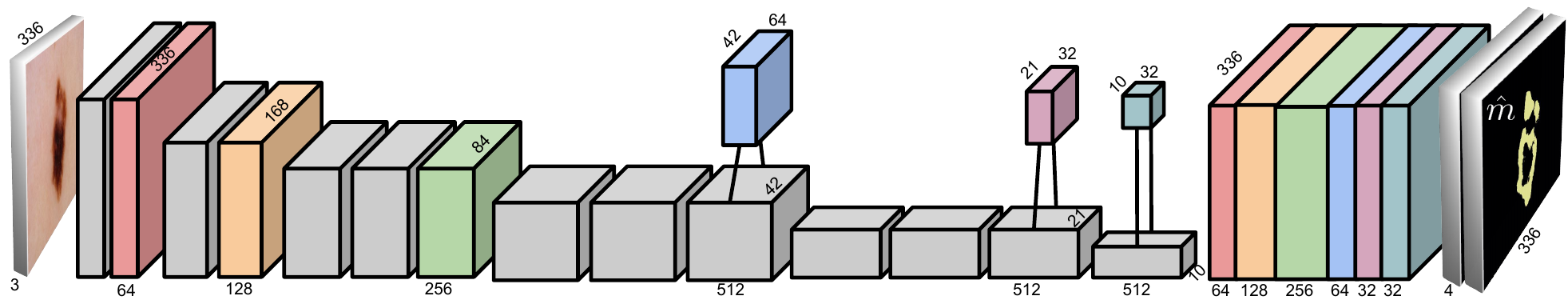}
\end{center}
\caption{The CNN used to segment clinical dermoscopic features. Feature maps from six layers are resized to match the spatial dimensions of the input and concatenated together. The colours indicate the selected layers that correspond to the concatenated block. We add additional convolutional layers to the deeper layers in order to reduce the number of feature maps (\emph{floating blocks}). A final layer is added to represent each of the dermoscopic features.}
\label{isic:fig:arch}
\end{figure*}

\subsection{Segmentations to Superpixels} While our CNN produces segmentations/pixel predictions (Fig.~\ref{fig:superpixelseg} \emph{c, f}), our final task is to assign a set of labels to each superpixel. We convert the predicted segmentation mask $\hat{m} \in  \mathbb{R}^{K \times W \times H}$ back to a predicted superpixel labelling $\hat{l}$ (Fig.~\ref{fig:superpixelseg} \emph{d, g}) by assigning to the $k$-th label of the $i$-th superpixel the average probabilities predicted within the $i$-th superpixel location, \ie
\begin{equation}
    \hat{l}_{ik} = \frac{1}{|s_i|} \sum_{w,h\in s_i} \hat{m}_{kwh}
\end{equation}
where $|s_i|$ indicates the number of pixels in the superpixel $s_i$, and $\hat{m}_{kwh}$ is the predicted probability of the $k$-th label at the $(w,h)$ spatial location.

\subsection{CNN Architecture} We extend VGG16~\cite{Simonyan2015}, a convolutional neural network, pretrained over ImageNet~\cite{Russakovsky2014}, using a similar semantic segmentation architecture as proposed by Long \etal~\cite{Long2015}. We remove the fully-connected layers of VGG16, and resize selected responses/feature maps throughout the network (see Fig.~\ref{isic:fig:arch} for selected layers) to match the sized of the input image using bilinear interpolation. These selected resized feature maps are concatenated, allowing us to directly consider feature maps from several network layers. This design is motivated by our observation that the appearance of clinical dermoscopic features are subtle, and may be represented in shallower layers with higher spatial resolutions. However, concatenating these resized responses from several layers results exceeds the memory available on modern GPUs. To lower the GPU memory requirements, and to give emphasis on feature maps from shallower layers, we reduce the number of concatenated feature maps from layers with 512 feature maps by adding additional convolutional layers with filters of size $512 \times 1 \times 1 \times F$, where $F$ is either 64 or 32 depending on the layer (Fig.~\ref{isic:fig:arch} provides details). This reduces GPU memory requirements, giving more emphasis to shallower layers, while still considering information found in deeper layers. Our final concatenated layer is of size $W \times H \times 576$, which matches the spatial dimensions of the input image $x$.

Our final layer adds an additional convolutional layer with a filter of size $576 \times 1\times 1 \times K$ to the concatenated block. This represents our output (\ie segmentation) for each of the $K$ dermoscopic features. A sigmoid activation function is applied element-wise to scale the output between 0 and 1. These $K$ additional \emph{channels} represent the labels for the $K$ types of dermoscopic features. Note that we do not apply the softmax activation function to this final layer, since dermoscopic clinical features can overlap.

\begin{figure}
\begin{center}
\includegraphics[width=0.98\linewidth]{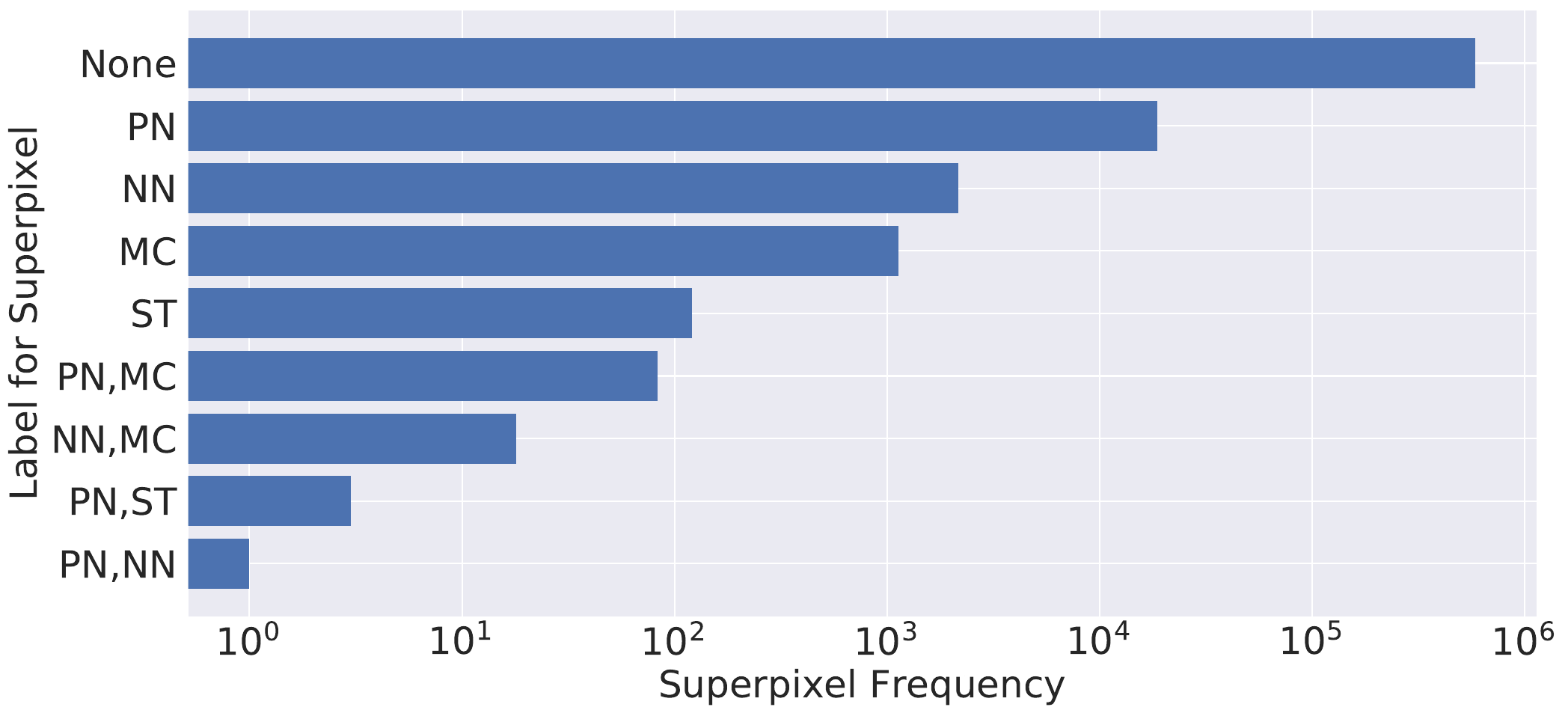}
\end{center}
\caption{The distribution of the superpixel labels over the ISIC-ISBI 2017 test set. The \emph{x-axis} shows the number of superpixels with a given label on a log scale, which illustrates the imbalanced data. The \emph{y-axis} shows the labels, and is expanded to show the frequency of superpixels that are assigned multiple labels. We see that most labeled superpixels have a single label (\eg pigment network \emph{PN} occurs most frequently on its own), but a single superpixel can contain multiple labels (\eg negative network \emph{NN} and milia-like \emph{MC} occur within the same superpixel). The majority of superpixels contain no label (\emph{None}). Some labels do not occur within the same superpixel (\eg streaks \emph{ST} never occurs with \emph{NN}) and are not shown here.}
\label{isic:fig:label_hist}
\end{figure}

\subsection{Negative Multi-Label S\o{}rensen-Dice-F$_1$ Loss Function} 
\label{isic:sec:loss}
The labels $l$ are heavily imbalanced in favour of the background, and even among the labels, some label types occur much more frequently than others. For example, in the \emph{ISIC-ISBI Part 2} challenge training data, there are approximately $55\times$ more pixels labelled as \term{pigment network}, than \term{negative network} (see Fig.~\ref{isic:fig:label_hist} for the distribution of labels). Additionally, many images contain no positive instances of a specific class. We consider data imbalance from three perspectives: \emph{pixel-imbalance}, where the background pixels dominate the foreground pixels; \emph{class-imbalance}, where some classes occur more frequently than others; and, \emph{sample-imbalance}, where many samples contain no positive instances. In order to encourage the CNN to be sensitive to clinical features and address pixel-imbalance, we base our loss on the S\o{}rensen-Dice-F$_1$ score. The F$_1$ score for two multi-dimensional arrays $\hat{a},a$ with $n$ elements, where $\hat{a}_i,a_i \in [0,1]$, is defined as
\begin{equation}
\label{isic:eq:f1}
    D(\hat{a}, a) = \frac{2 \cdot TP(\hat{a}, a) + \alpha}{2 \cdot TP(\hat{a}, a) + FP(\hat{a}, a) + FN(\hat{a},a) + \beta}
\end{equation}
where fuzzy true positives $TP(\hat{a},a) = \sum_{i}^n (\hat{a}_i \cdot a_i)$, false positives $FP(\hat{a},a) = \sum_i^n(\hat{a}_{i} \cdot (1-a_i))$, and false negatives $FN(\hat{a},a) = \sum_i^n((1-\hat{a}_i) \cdot a_i)$ are computed~\cite{Pastor-Pellicer2013}. Setting $\beta > 0$ prevents divide-by-zero errors and $\alpha$ controls the score returned when neither the ground truth nor the predicted labels have any positive values. Equation~\ref{isic:eq:f1} can be simplified and rewritten into an equivalent form more recognizable as the S\o{}rensen-Dice coefficient,
\begin{equation}
    D(\hat{a},a) = \frac{2 \cdot \sum_i^n (\hat{a}_i \cdot a_i) + \alpha}{\sum_i^n(\hat{a}_i + a_i) + \beta} ~.
\end{equation}

The loss function to train a CNN is computed over mini-batches $\hat{M} \in \mathbb{R}^{B \times K \times W \times H}$, where $B$ is the number of mini-batch samples (\eg $\hat{m}$ is a single sample). Given the predicted $\hat{M}$ and true $M$ mini-batch segmentations, we train the CNN to minimize a negative multi-label S\o{}rensen-Dice-F$_1$ score
\begin{equation}
\label{isic:eq:loss}
     \ell(\hat{M},M) = 1 - D^*(\hat{M}, M)
\end{equation}
where $D^*(\hat{M}, M)$ computes the S\o{}rensen-Dice-F$_1$ score over a mini-batch. $D^*(\hat{M}, M)$ can take different forms by computing $D(\cdot, \cdot)$ over various mini-batch partitions. For example, if $D^*(\hat{M}, M) = D(\hat{M}, M)$, we compute a \emph{single} S\o{}rensen-Dice-F$_1$ score for the entire mini-batch, which addresses pixel-imbalance. However, class-imbalance can cause the model to be biased towards the prevalent class label, which can result in the model ignoring infrequent class labels. To balance infrequent class labels, an intuitive choice which avoids explicit class re-weighting (as in~\cite{Sudre2017}) is to compute the S\o{}rensen-Dice-F$_1$ score over each of the $K$ channels, and over each of the $B$ mini-batch samples,
\begin{equation}
\label{isic:eq:channel-image-batch}
    D^{B,K}(\hat{M}, M) = \frac{1}{B\cdot K} \sum_b^B \sum_k^K D(\hat{M}^{b,k,:,:}, M^{b,k,:,:})
\end{equation}
where $M^{b,k,:,:}$ represents a 2D array that corresponds to the $b$-th sample of the $k$-th channel. Setting $\alpha, \beta = 1$ avoids divide by zero errors, and returns a score of 1 when both the predicted and ground truth labels are all zeros (loss = 0 Eq.~\ref{isic:eq:loss}). However, in datasets where a large proportion of samples contain no positive labels (\ie sample-imbalance), this can bias the classifier to learn to only predict background labels. Setting $\alpha=0$ and $\beta=1$ returns a score of 0 (loss = 1) when both the predicted and ground truth are all zero. While this no longer encourages the model to learn to predict all background values, it considers all negative samples as an error regardless of the predictions, which prevents the model from learning using the negative samples. In order for the model to learn from negative samples, and to account for sample and class-imbalance without explicit re-weighting, for each channel, we compute a S\o{}rensen-Dice-F$_1$ score over the entire $B$ samples within the mini-batch,  
\begin{equation}
\label{isic:eq:channel-batch}
    D^K(\hat{M}, M) = \frac{1}{K}  \sum_k^K D(\hat{M}^{:,k,:,:}, M^{:,k,:,:})
\end{equation}
where $M^{:,k,:,:}$ represents a 3D array composed of the $k$-th channel of all $B$ samples within a mini-batch. Cases when the entire ground truth channel is composed of all negative samples will occur less frequently since $B$ samples are considered simultaneously. Thus, computing the S\o{}rensen-Dice-F$_1$ score for each mini-batch channel (rather than for each sample) allows negative samples to contribute to the learning without dominating the loss function.

\subsection{Training and Augmented Data with Over-Sampled Classes}
We train our CNN by minimizing Eq.~\ref{isic:eq:f1} using the Adam optimizer~\cite{Kingma2015} with a learning rate of 0.00005. Our models were built and optimized using Keras~\cite{chollet2015keras} with TensorFlow~\cite{tensorflow2015-whitepaper}. While VGG is trained on images of size $224\times224$ for classification, we use larger image resolutions of $336 \times 336$, which is possible since all our layers are convolutional. We use a mini-batch of size 12 as larger batches exceeded our GPU memory. We apply real time data augmentation, where in each mini-batch, the data is augmented (\eg flips, rotations) and the mini-batch is randomly sampled such that at least two samples contain each of the class labels. The remaining four are randomly sampled. For our ISIC-ISBI entry, we did not use data augmentation nor over-sampling, and stopped training after only 5 epochs, as empirically we found longer training yielded segmentations less sensitive to the clinical features. For our subsequent experiments, we show experiments with and without data augmentation/over-sampling, train for 100 epochs, and choose the model that achieves the lowest loss over our validation set.

\section{Results and Discussions}
We trained our network over 1700 images from the ISIC-ISBI 2017 skin analysis challenge, and used 300 images to monitor the network's performance with different hyperparameters. The public leaderboard consisted of 150 images, with a separate private leaderboard of 600 images. While several metrics were evaluated, the winner of the challenge was determined by the highest averaged Area Under the Receiver Operator Characteristic curve (AUROC). Our approach achieved the highest averaged AUROC when compared to the other entries. The results over both the public validation and private test sets were fairly consistent. The results for ours and competing approaches over the private test set of 600 images are shown in Table.~\ref{isic:tab:results}. We composed Table~\ref{isic:tab:results} from the online submission system~\cite{isic-isbi-part2}, which was evaluated over a controlled submission server and only made public after the competition.

\begin{table}[ht]
    \caption{Official results over the ISIC-ISBI 2017 test dataset.}
    \centering
    \begin{tabular}{@{~}c@{~}|@{~}c@{~}|c|@{~}c@{~}|c|c|c@{~}}
Entry & Dermoscopic Feature & ACC & AUROC & AP   & SEN  & SPC \\
\hline

Lee  & pigment network   & 0.915 & 0.828 & 0.487          & 0.736 & 0.921 \\
\cite{Li2017} & negative network   & 0.905 & 0.762 & \textbf{0.321} & \textbf{0.618} & 0.906 \\
     & milia-like cysts  & 0.843 & 0.837 & \textbf{0.421} & \textbf{0.832} & 0.843 \\
     & streaks  & 0.961 & 0.900 & \textbf{0.422} & \textbf{0.839} & 0.961 \\
     & average & 0.906 & 0.832 & \textbf{0.413} & 0.649 & 0.907 \\
\hline

Shen & pigment network   & 0.909 & 0.835 & 0.491 & 0.756 & 0.914 \\
\cite{Li2017}   & negative network   & 0.917 & 0.762 & 0.317 & 0.606 & 0.919 \\
     & milia-like cysts  & 0.852 & \textbf{0.838} & 0.418 & 0.824 & 0.852 \\
     & streaks  & 0.978 & 0.896 & 0.411 & 0.815 & 0.978 \\
     & average & 0.914 & 0.833 & 0.409 & \textbf{0.665} & 0.915 \\
\hline

ours & pigment network   & \textbf{0.951} & \textbf{0.945} & \textbf{0.582} & \textbf{0.803} & \textbf{0.956} \\
     & negative network   & \textbf{0.982} & \textbf{0.869} & 0.152          & 0.428          & \textbf{0.984} \\
     & milia-like cysts  & \textbf{0.988} & 0.807          & 0.078          & 0.303          & \textbf{0.990} \\
     & streaks  & \textbf{0.997} & \textbf{0.960} & 0.151          & 0.637          & \textbf{0.997} \\
     & average & \textbf{0.980} & \textbf{0.895} & 0.241          & 0.542          & \textbf{0.981} \\
\hline
    \end{tabular}
    \label{isic:tab:results}
    \\[5pt]
    \justify
    Results are divided by challenge entry and dermoscopic feature type. The \emph{average} row averages results over all features in the dataset. \emph{ACC} represents accuracy, \emph{AP} represents average precision, \emph{SEN} represents sensitivity, \emph{SPC} represents specificity.
\end{table}

\subsection{Dermoscopic Feature Classification - Challenge Results}
From Table~\ref{isic:tab:results}, we observe the challenges and importance of choosing appropriate metrics when evaluating different methods. In addition to the metric of AUROC, accuracy, average precision, sensitivity, and specificity, were also evaluated. While AUROC was chosen as single metric to rank entries, and our approach achieved higher AUROC when compared to the other entries (ours 0.895 vs second place 0.833~\cite{Li2017}), the other entries outperform our approach on other metrics. 

As the entry by Li and Shen~\cite{Li2017} is a superpixel classification approach using a CNN, evaluated over the same dataset, we can compare superpixel classification with our semantic segmentation approach. In general, we see that our approach is less sensitive, but more specific when detecting dermoscopic features. Notably, for the \term{pigment network} dermoscopic feature, we achieve the highest results across all metrics.

\begin{figure*}
\begin{center}
\includegraphics[width=0.98\linewidth]{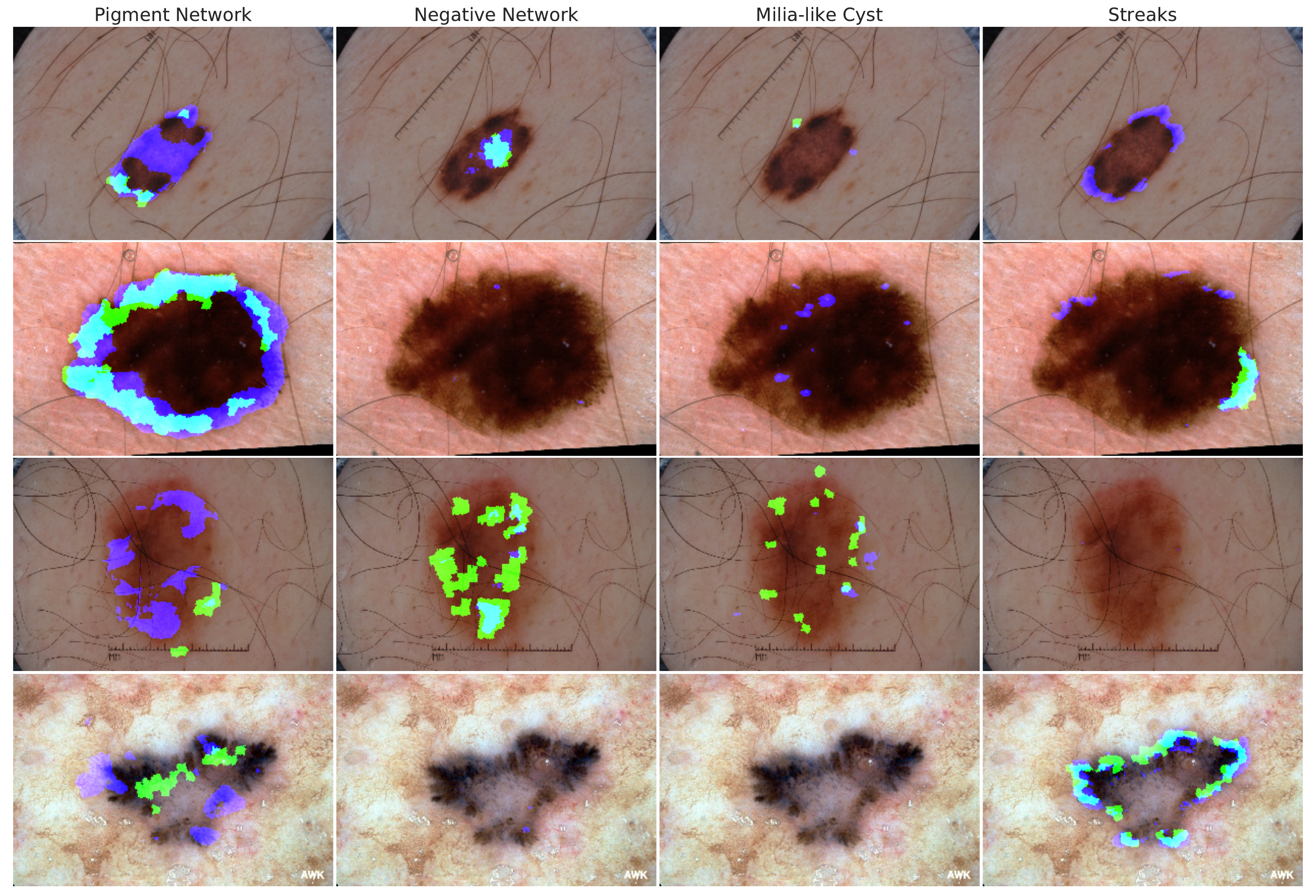}
\end{center}
\caption{Dermoscopic features overlaid on the skin images. Each type of clinical dermoscopic feature (columns) is overlaid on four sample images from the test set (rows). \emph{Green} pixels indicate ground truth. \emph{Dark blue} pixels represent pixels predicted to have the specific feature. \emph{Light blue} pixels indicate an overlap between predicted and ground truth.}
\label{isic:fig:derm-features}
\end{figure*}

We show example results of the predicted and ground truth pixels for each type of clinical dermoscopic feature in Fig.~\ref{isic:fig:derm-features}. This figure highlights the challenges of detecting dermoscopic features, as the visual cues for the various clinical features are subtle and often not obvious to an untrained eye. We observe that \term{pigment network} and \term{streaks} often occur near the boundary of the lesion, while \term{negative network} can occur within the lesion. This illustrates how the context of the superpixel (\ie information in surrounding pixels) is an important factor to consider when detecting dermoscopic features, and supports our approach to frame this task as segmentation problem, rather than classifying individual superpixels.

\begin{table}[ht]
    \caption{Two simple baselines experiments.}
    \centering
    \begin{tabular}{@{~}c@{~}|@{~}c@{~}|@{~}c@{~}|@{~}c@{~}|c|c|c|c|c@{~}}
Exp. & DCF & ACC & AUROC & AP   & SEN  & SPC & $\bar{J}_1$ & $\bar{J}_{\mathrm{nan}}$ \\
\hline
Lesion
 & PN & 0.832 & 0.913 & 0.528 & 0.962 & 0.827 & 0.167 & 0.167 \\
 & NN & 0.807 & 0.916 & 0.502 & 0.992 & 0.806 & 0.012 & 0.012 \\
 & MC & 0.805 & 0.884 & 0.421 & 0.915 & 0.805 & 0.016 & 0.016 \\
 & ST & 0.803 & 0.894 & 0.380 & 0.960 & 0.803 & 0.001 & 0.001 \\
 & avg & 0.812 & \textbf{0.902} & 0.458 & \textbf{0.957} & 0.810 & 0.049 & 0.049 \\

\hline
Empty
 & PN & 0.969 & 0.500 & 0.515 & 0.000 & 1.000 & 0.445 & 0.000 \\
 & NN & 0.996 & 0.500 & 0.502 & 0.000 & 1.000 & 0.925 & 0.000 \\
 & MC & 0.998 & 0.500 & 0.501 & 0.000 & 1.000 & 0.755 & 0.000 \\
 & ST & 1.000 & 0.500 & 0.500 & 0.000 & 1.000 & 0.985 & 0.000 \\
 & avg & \textbf{0.991} & 0.500 & \textbf{0.505} & 0.000 & \textbf{1.000} & \textbf{0.777} & 0.000 \\

\hline
ours$^*$
 & PN & 0.951 & 0.944 & 0.585 & 0.806 & 0.956 & 0.319 & 0.217 \\
(\emph{ISIC} & NN & 0.982 & 0.870 & 0.159 & 0.427 & 0.984 & 0.339 & 0.021 \\
\emph{entry})  & MC & 0.988 & 0.809 & 0.075 & 0.294 & 0.990 & 0.225 & 0.031 \\
 & ST & 0.997 & 0.963 & 0.154 & 0.605 & 0.997 & 0.532 & 0.007 \\
 & avg & 0.980 & 0.896 & 0.243 & 0.533 & 0.982 & 0.354 & \textbf{0.069} \\

\hline
    \end{tabular}
    \\[5pt]
    \justify
    \emph{Lesion} indicates that the predicted lesion segmentation is used for all dermoscopic features predictions. \emph{Empty} indicates that only background is predicted. \emph{DCF} is short for dermoscopic clinical feature. $J_1$ and $J_{\mathrm{nan}}$ represent the Jaccard Index with different values assigned to the empty set. Over the ISIC-ISBI 2017 test dataset, these simple baselines outperform existing methods when ranked using the challenge metrics, but not when ranked using the $J_{\mathrm{nan}}$ metric. $^*$We report slight ($\approx$1\%) differences from the official results in Table~\ref{isic:tab:results}.
    \label{isic:tab:compare}
\end{table}

\subsection{Dermoscopic Feature Classification - Simple Baselines}
We show that two simple baseline approaches (Table~\ref{isic:tab:compare} experiments \emph{Lesion} and \emph{Empty}) outperform existing methods when ranked using the metrics from Part 2 of the ISIC-ISBI 2016~\cite{Gutman2016} and 2017~\cite{Codella2017a} challenge. For the first baseline approach (Table~\ref{isic:tab:compare} Exp. \emph{Lesion}), we use a trained lesion segmentation model (described in Sec.~\ref{isic:sec:lesion-seg}) to label all pixels within a predicted lesion segmentation mask as positive incidences of each dermoscopic feature. Surprisingly, this simple approach achieves the highest averaged AUROC (used to rank Part 2 of the 2017 challenge~\cite{Codella2017a}) and average precision score (used to rank Part 2A of the 2016 challenge~\cite{Gutman2016}), outperforming existing methods (Table~\ref{isic:tab:compare} Exp. \emph{Lesion}). Although this approach scores high on the official benchmarks, considering the entire lesion as a clinical dermoscopic feature is not practically useful. In order to establish a metric that better captures the utility of the results, we propose to use a fuzzy Jaccard Index~\cite{Crum2006}, defined as,
\begin{equation}
    J(\hat{a}, a) 
    = f_{\mathrm{nan}}\left( \frac{\sum_i^n \mathrm{min}(\hat{a}_i, a_i)}
    {\sum_i^n \mathrm{max}(\hat{a}_i, a_i)}
    \right)
\end{equation}
where the $\mathrm{min}(\cdot, \cdot)$ and $\mathrm{max}(\cdot, \cdot)$ functions compute a probabilistic intersection and union, respectively; $f_{\mathrm{nan}}(x) = $ \emph{nan} if the denominator is 0 else $x$; and \emph{nan} is a sentinel indicating an undefined value. Given a test set of $N$ predicted $\hat{M} \in \mathbb{R}^{N \times K \times W \times H}$ and ground truth $M$ segmentations, computing the Jaccard Index over the entire  (\ie $J(\hat{M}, M)$), will bias results towards more frequently occurring classes. Computing the Jaccard Index for each channel separately, $J_c(\hat{M}^{:,k,:,:}, M^{:,k,:,:})$ (this is how Part 2B~\cite{Gutman2016} appeared to be ranked), will reduce the contribution of images with a relatively small proportion of positive pixels. In order to give higher weight to images with smaller dermoscopic features, we average over each image, 
\begin{equation}
    \bar{J}_1(\hat{M}^{:,k,:,:},M^{:,k,:,:}) = \frac{1}{N} \sum_i^N f_1(J(\hat{M}^{i,k,:,:}, M^{i,k,:,:}))
\end{equation}
where $\hat{M}^{:,k,:,:}$ are all $N$ predictions for the $k$-th channel. An intuitive function that considers \emph{nan} values, is to let $f_1(x) = 1$ if $x =$ \emph{nan}, else $x$, which returns a Jaccard Index of 1 when there are neither any positive predicted nor ground truth cases (\ie the empty set). Using this measure, our proposed approach (Table~\ref{isic:tab:compare} Exp. \emph{ours}) scores considerably higher than the \emph{Lesion} experiment, suggesting that $\bar{J}_1$ is a more informative metric than AUROC or the average precision score. However, in imbalanced datasets where many images contain no positive labels (Fig.~\ref{isic:fig:label_hist}), a classifier that predicts \emph{only} background can achieve a high score. We empirically show that by predicting only background (Table~\ref{isic:tab:compare} Exp. \emph{Empty}), we achieve a higher Jaccard Index. Thus, we propose 
\begin{equation}
    \bar{J}_\mathrm{nan}(\hat{M}^{:,k,:,:},M^{:,k,:,:}) = \frac{\sum_i^N f_0( J(\hat{M}^{i,k,:,:}, M^{i,k,:,:}) ) }{\sum_i^N f_{01}(J(\hat{M}^{i,k,:,:}, M^{i,k,:,:}) } 
\end{equation}
where $f_0(x) = 0$ if $x =$ \emph{nan}, else $x$ and $f_{01}(x) = 0$ if $x=$ \emph{nan} else $1$. This excludes all images where both the predicted and ground truth do not include any positive samples. $\bar{J}_{\mathrm{nan}}$ penalizes a model that only assigns a background label (Exp. \emph{Empty}), and our approach (Exp. \emph{ours}) produces consistently higher $\bar{J}_{\mathrm{nan}}$ scores than the \emph{Lesion} experiment. We note that when computing the Jaccard Index, rather than using the predictions $m$ directly, we use the superpixel probabilities (\eg~Fig.~\ref{fig:superpixelseg} \emph{d,g}), \ie use $\hat{l}_{ik}$ in Eq.~\ref{isic:eq:super_to_seg}, where $\hat{l}_{ik} = 0$ if $\hat{l}_{ik} < 0.5$ else $\hat{l}_{ik}$. This is done to remove false positive superpixels. Quantitative results showing of averaged improvements after thresholding and converting to superpixel segmentation are given in Table~\ref{isic:tab:modelvariants}.

\subsection{Lesion Segmentation}
\label{isic:sec:lesion-seg}
\begin{figure}
\begin{center}
\includegraphics[trim={0 11.1cm 0 0},clip, width=0.98\linewidth]{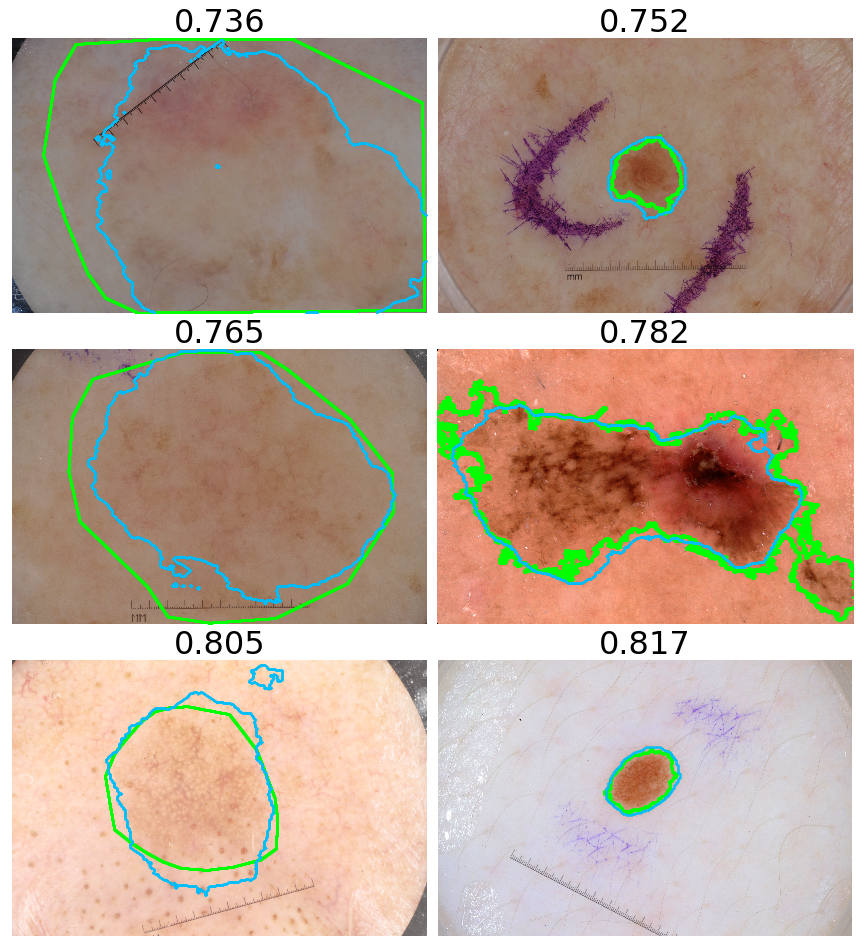}
\end{center}
\vspace{-1em}
\caption{Example segmentation results where the \emph{green line} indicates the ground truth contour, and the \emph{blue line} represents our predicted lesion contour. The Jaccard Index between the predicted and ground truth lesion are displayed above each image. These cases illustrate where the exact lesion borders may be subjective. Note the variability in the ground truth borders (\eg some have straight lines, while others are highly sensitive to intensity changes).}
\label{isic:fig:segment}
\end{figure}

While not a focus of this paper, we note that our entry for \emph{Part 1: Lesion Segmentation Task} ranked sixth out of 21 entries based on the Jaccard distance (ours 0.752 vs first place 0.765~\cite{Yuan2017}). For our segmentation entry, we used nearly the same model and loss as described in this paper. Notable differences include: images were resized $224 \times 224$; the original feature maps were used from the deeper layers; an additional convolutional layer after the concatenated layer was added; and, the model was trained for 12 epochs with a batch size of eight. Our competitive results over the segmentation challenge using only minor modifications suggests both lesion segmentation (Part 1) and dermoscopic clinical feature detection (Part 2) can be approached in similar ways. Fig.~\ref{isic:fig:segment} shows examples where the contours of the ground truth and the predicted lesions are overlaid on the original lesion images. We sampled lesions that have a computed Jaccard Index around the range of the top performing methods (sampled between 0.736 and 0.782 Jaccard Index), to show the variability and subjectivity of the lesion borders in certain cases. Given the subjectivity observed in defining precise lesion borders, and the similarity between the top performing approach~\cite{Yuan2017} and ours (only a 0.013 Jaccard Index difference), our segmentation approach is competitive with current state-of-the-art methods.

\begin{table}[ht]
    \caption{Detailed results comparing loss functions.}
    \centering
    \begin{tabular}{c|c|c|c|c|c}
Exp. & DCF & AUROC & AP & $J_{\mathrm{c}}$ & $\bar{J}_{\mathrm{nan}}$ \\
\hline
                       &   PN &  0.963 &  0.578 &  0.299 &                     0.209 \\
            \emph{(a)} &   NN &  0.941 &  0.091 &  0.066 &                     0.027 \\
  \emph{Cross-entropy} &   MC &  0.948 &  0.077 &  0.037 &                     0.023 \\
 \emph{class-weighted} &   ST &  0.966 &  0.049 &  0.027 &                     0.009 \\
                       &  avg &  \textbf{0.955} &  0.199 &  0.107 &                     0.067 \\
\hline
                          &   PN &  0.882 &  0.591 &  0.427 &                     0.269 \\
               \emph{(b)} &   NN &  0.502 &  0.008 &  0.000 &                     0.000 \\
           \emph{Dice-F1} &   MC &  0.500 &  0.001 &  0.000 &                     0.000 \\
 \emph{volume-mini-batch} &   ST &  0.500 &  0.000 &  0.000 &                     0.000 \\
Eq.~\ref{isic:eq:f1}  &  avg &  0.596 &  0.150 &  0.107 &                     0.067 \\

\hline

                      &   PN &  0.938 &  0.591 &  0.380 &                     0.232 \\
           \emph{(c)} &   NN &  0.798 &  0.113 &  0.080 &                     0.027 \\
       \emph{Dice-F1} &   MC &  0.793 &  0.075 &  0.094 &                     0.045 \\
 \emph{channel-image} &   ST &  0.845 &  0.033 &  0.046 &                     0.010 \\
Eq.~\ref{isic:eq:channel-image-batch} &  avg &  0.843 &  0.203 &  0.150 &                     0.078 \\
\hline
                      &   PN &  0.910 &  0.602 &  0.426 &                     0.282 \\
           \emph{(d)} &   NN &  0.645 &  0.134 &  0.079 &                     0.056 \\
       \emph{Dice-F1} &   MC &  0.737 &  0.103 &  0.126 &                     0.051 \\
 \emph{channel-batch} &   ST &  0.641 &  0.053 &  0.048 &                     0.039 \\
 Eq.~\ref{isic:eq:channel-batch}  &  avg &  0.733 &  \textbf{0.223} &  \textbf{0.170} & \textbf{0.107} \\
          
\hline
    \end{tabular}
    \label{isic:tab:detailedvariants}
    \\[5pt]
    \justify
    The \emph{cross-entropy} loss is weighted to account for class imbalance. We display the ranking metrics, and note that while experiment \emph{(a)} achieves the highest AUROC, we propose that the Jaccard Index $\bar{J}_{\mathrm{nan}}$ better quantifies the performance of a model at the intended task.
\end{table}

\subsection{Comparing Losses and Model Variants}
We compare the Dice-F1 loss function with a weighted binary cross-entropy loss function, where we weight each pixel using median frequency balancing~\cite{Eigen2015}. Using the weighted binary cross-entropy loss averaged over the four dermoscopic features as our loss function, the model converges to predicting \emph{all} background labels (Table~\ref{isic:tab:lossvariants} - first row). Oversampling the minority class during data-augmentation improves results (Table~\ref{isic:tab:detailedvariants} \emph{a}). While the resulting AUROC curve is higher than previously reported, the computed Jaccard Index is relatively low, indicating an over-segmentation similar to using the predicted lesion (Table~\ref{isic:tab:compare} - \emph{Lesion}).

Our subsequent experiments compare different mini-batch partitions when computing the Dice-F1 score. When computing the Dice-F1 score over the entire mini-batch over all labels (\ie $D^*=D$ Eq.~\ref{isic:eq:loss}), only the larger \emph{pigment network} class performs well (Table~\ref{isic:tab:detailedvariants}~\emph{b}). Averaging the loss over each mini-batch sample, over each label-channel (Eq.~\ref{isic:eq:channel-image-batch} $D^*=D^{B,K}$) further improved results (Table~\ref{isic:tab:detailedvariants}~\emph{c}). Computing the Dice-F1 score over the entire channel within a mini-batch (Eq.~\ref{isic:eq:channel-batch} $D^*=D^{K}$), yields the top Jaccard Index (Table~\ref{isic:tab:detailedvariants}~\emph{d}).

In Table~\ref{isic:tab:lossvariants}, we show the model performance with setting $\alpha, \beta$ in Eq.~\ref{isic:eq:f1} and through class oversampling during data augmentation. The cases where the model converges to predicting all background ($\bar{J}_{\mathrm{nan}}$=0) indicates the challenges with infrequent class labels within imbalanced datasets.

We also experiment with substituting VGG16 with more recent models: ResNet50~\cite{He2016}, and InceptionResNetV2~\cite{Szegedy2017}.  We find that changing the underlying model did not improve results. We suspect VGG is particularly well suited to this task since the first two convolutional layers of VGG16 maintain the original spatial dimensions of the input, producing high resolution feature maps that are directly considered in the output segmentation layer (in contrast ResNet50 reduces the spatial dimension in half after the first convolutional layer). As the clinical dermoscopic features occupy only a  fraction of the entire image, these high resolution feature maps may be necessary to detect subtle image cues.

Our final experiment replaces the concatenated skip connections with UNet~\cite{Ronneberger2015} connections. This did not improve the final result after thresholding and converting to superpixels. This may in part be due to the increased number of parameters that need to be learned to incorporate deeper feature maps. While these more recent models and modifications to the architecture did not improve results, we highlight that the Dice-F1 loss function is not model specific, and other segmentation models may yield further improvements.

\begin{table}[ht]
    \caption{Experiments computing the loss over different mini-batch partitions and correcting for divide-by-zero errors.}
    \centering
    \begin{tabular}{@{~}c|c|c|c|c|c|c@{~}}
    Loss & \multicolumn{2}{|c|}{Compute over} & Class-augment & $\alpha$ & $\beta$ & $\bar{J}_{\mathrm{nan}}$ \\
    \hline
    Cross-entropy & - & - & No & - & - & 0.0 \\
    Cross-entropy & - & - & Yes & - & - & 0.067 \\
    Dice-F1 & Volume & Batch & Yes & 0 & 1 & 0.067 \\
    Dice-F1 & Channel & Image & Yes & 1 & 1 & 0.0 \\
    Dice-F1 & Channel & Image & Yes & 0 & 1 & 0.078 \\
    Dice-F1 & Channel & Batch & No & 1 & 1 & 0.0 \\
    Dice-F1 & Channel & Batch & No & 0 & 1 & 0.083 \\
    Dice-F1 & Channel & Batch & Yes & 0 & 1 & 0.107 \\
    \hline
    \end{tabular}
    \label{isic:tab:lossvariants}
    \\[5pt]
    \justify
    These results highlight the importance of choosing the appropriate mini-batch partition, and how subtle differences in correcting for divide-by-zero errors, or improper class weighting, can yield a model that converges to predicting all background values (denoted as $\bar{J}_{\mathrm{nan}} = 0$).
\end{table}

\begin{table}[ht]
    \caption{Base models and segmentation connection types experiments.}
    \centering
    \begin{tabular}{@{~}c@{~}|@{~}c@{~}|@{~}c@{~}|@{~}c@{~}|@{~}c@{~}}
Base-model & Type &  Direct-$\bar{J}_{\mathrm{nan}}$ &  Thresh-$\bar{J}_{\mathrm{nan}}$ &  $\bar{J}_{\mathrm{nan}}$ \\
\hline
    InceptionResNetV2~\cite{Szegedy2017} & Skip~\cite{Long2015} & 0.045 & 0.078 & 0.082 \\
    ResNet50~\cite{He2016} & Skip & 0.049 & 0.083 & 0.091 \\
    VGG~\cite{Simonyan2015} & UNet~\cite{Ronneberger2015} & 0.072 & 0.073 & 0.082 \\
    VGG & Skip & 0.071 & 0.088 & 0.107 \\
\hline
    \end{tabular}
    \label{isic:tab:modelvariants}
    \\[5pt]
    \justify
    Using VGG as a base model with concatenated \emph{skip} connections yielded slightly high averaged Jaccard Index results than other models and UNet type connections. This table also shows the results after using the direct prediction (\emph{Direct-}$\bar{J}_{\mathrm{nan}}$), after thesholding the predictions (\emph{Thresh-}$\bar{J}_{\mathrm{nan}}$), and converting the predictions to superpixels ($\bar{J}_{\mathrm{nan}}$).
\end{table}

\section{Conclusions}
Our method approached the superpixel labelling task as a segmentation problem, used a CNN architecture that relied on interpolated and concatenated feature maps from the intermediate network layers, and minimized a negative multi-label S\o{}rensen-Dice coefficient (F$_1$ score) computed across a partition of the mini-batch. We ranked first place in the ISIC-ISBI Part 2 Challenge, achieving the highest averaged area under the receiver operator characteristic curve over both the public validation and private test-set leaderboard. For individual dermoscopic features, we had the highest AUROC score for pigment network, negative network, and streaks. We demonstrated how simple baseline methods rank higher than existing approaches when using the current ranking metrics, and propose to use the averaged fuzzy Jaccard Index that ignores the values of the empty set. We highlight that the very low results reported using the averaged Jaccard Index from our top performing model (0.107), indicates significant room for improvement in this task, which is not as obvious when reporting the high (0.896) AUROC score. The ability to detect \term{pigment network} within dermoscopic images shows promise, although the low average precision and Jaccard Index indicates this task can be greatly improved. The low performance detecting other clinical dermoscopic features remains an area for future research. Our competitive results over the Part 1 Segmentation challenge using nearly the same method, suggests both segmentation and clinical feature detection can be approached in similar ways. We hope the release of our code and trained model will serve as a baseline approach on which other groups can improve.

\subsubsection*{Acknowledgments}
The authors are grateful to Kathleen P. Moriarty for helpful discussions and assistance in data preparation, and to the NVIDIA Corporation for donating a Titan X GPU used in this research. 

\bibliographystyle{IEEEtran}
\bibliography{IEEEabrv,refs}{}

\end{document}